\documentclass[letterpaper, 10 pt, conference]{ieeeconf}  %

\IEEEoverridecommandlockouts                              %

\usepackage{amsmath} %
\usepackage{amssymb}  %
\usepackage{multirow}
\usepackage{multicol}
\usepackage{booktabs}
\usepackage{graphicx}
\usepackage{array}

\usepackage{capt-of}
\usepackage{hyperref}

\title{\LARGE \bf
Real-Time Force Regulation for Whole-Hand Dexterous Grasping
}

\author{%
Sang Min Kim$^{1\dagger}$,
Alexander Alexiev$^{2}$,
Tzu-Yuan Lin$^{2}$,
Sangbae Kim$^{2}$,
Young Min Kim$^{1*}$, and
Yonghyeon Lee$^{3*}$%
\thanks{%
$^{1}$Department of Electrical and Computer Engineering,
Seoul National University, Seoul, Republic of Korea.
$^{2}$Department of Mechanical Engineering,
Massachusetts Institute of Technology, Cambridge, MA, USA.
$^{3}$Department of Artificial Intelligence,
Yonsei University, Seoul, Republic of Korea.
$^{\dagger}$Work done while Sang Min Kim was visiting MIT.
$^{*}$Co-corresponding authors:
\texttt{youngmin.kim@snu.ac.kr} and
\texttt{yonghyeon.lee@yonsei.ac.kr}.}%
}

\IEEEaftertitletext{
  \begin{center}
    \includegraphics[width=\textwidth]{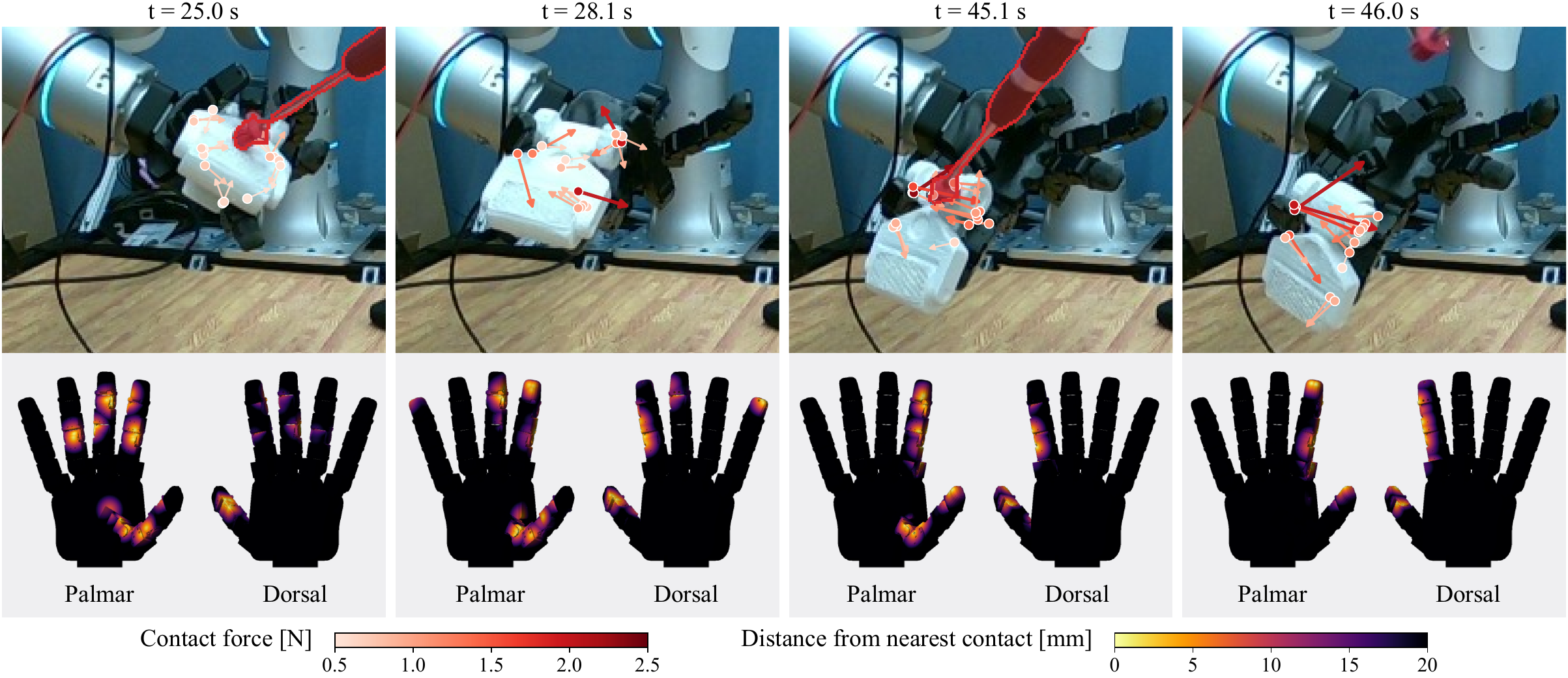}
    \captionof{figure}{
Real-time force regulation over whole-hand contacts on a 27-DoF arm--hand system under external perturbations.
Top: snapshots of the experiment.
A human perturbs the object with the rod highlighted in red.
Arrows show the commanded contact forces ($\geq 0.5$\,N) at the detected contact points, colored by the magnitude.
Bottom: the Robotis 5F hand in an open pose, colored by distance to the nearest contact.
Note that contacts are detected not only at the fingertips but also on the sides and back of the fingers (see the dorsal rendering).
}
    \label{fig:realworld_overview}
  \end{center}
  \vspace{0em}
}

\begin{document}

\maketitle
\thispagestyle{empty}
\pagestyle{empty}

\begin{abstract}
Robust dexterous grasping requires maintaining physical stability despite contacts interactively evolving across the entire hand.
A precomputed force distribution can easily fail under object motion, modeling errors, or external disturbances.
In this paper, we present a framework for \textit{real-time} force regulation over dynamically changing whole-hand contacts.
Our method geometrically estimates contacts \textit{across all hand links} using a tracked object model and proprioception, without requiring tactile sensing at those contacts.
It repeatedly recomputes the desired contact-force distribution subject to friction constraints, actuator limits, and an actuation-consistency constraint motivated by classical whole-limb force analysis.
We integrate this force-regulation controller with reactive reaching, enabling the hand to acquire a grasp, maintain it under disturbances, and regrasp after losing the object.
Simulation experiments without gravity demonstrate improved grasp retention over fixed-allocation and fingertip-only execution under controlled perturbations, while real-world experiments on a 27-DoF arm--hand system demonstrate grasp maintenance and recovery under human-applied disturbances as contacts evolve across the whole hand. Project page: \url{https://sangminkim-99.github.io/reactive-grasp-whole-hand/}.
\end{abstract}

\section{Introduction}
Robust dexterous grasping is not only a grasp-pose generation problem, but also a problem of maintaining a stable physical interaction over time.
A planned grasp pose is hard to reproduce exactly, because of pose error, control error, and unplanned collisions along the way.
Even when it is reached, the fingers push the object as they close.
After the grasp, disturbances, collisions, and arm acceleration move it again.
The contacts thus differ from the plan from the start and keep shifting, appearing, and disappearing across the whole hand.
A precomputed force distribution therefore quickly becomes invalid: changes in contact locations and normals alter both the grasp matrix mapping contact forces to object wrenches and the contact Jacobian mapping them to joint torques.
Maintaining a feasible contact-force distribution thus requires continuously estimating contacts and reallocating forces as the contact set evolves.

Existing work addresses this problem only partially. Dexterous grasp-synthesis methods generate diverse and stable grasp poses for complex objects~\cite{wang2022dexgraspnet,weng2024dexdiffuser,zhang2024dexgraspnet,zurbrugg2025graspqp,chen2025dexonomy}. The resulting grasps, however, are typically executed open-loop: the fingers move to the planned configuration and apply precomputed forces or close further by a fixed amount. Although some reactive grasp controllers adapt the grasp online after contact~\cite{wimboeck2006passivity,khadivar2023adaptive,gold2021model,psomopoulou2021robust,ke2026tacdexgrasp,yu2026coorgrasp}, most consider only fingertip contacts, where tactile sensors are typically located. Extending such regulation to uninstrumented surfaces across the whole hand remains challenging.

An ideal grasp should exploit contacts across the entire hand, including not only the fingertips but also the finger sides, proximal links, and palm. Regulating these contacts requires estimating their locations and surface normals, yet existing methods provide only partial information. Tactile skins increasingly cover the phalanges and palm~\cite{zhao2025embedding,sharma2025self,zhang2025soft}, but instrumenting the full surface of an articulated hand remains challenging~\cite{yousef2011tactile,slepyan2026scalable}. Joint-torque-based methods can localize a single contact but become ambiguous when multiple contacts occur simultaneously~\cite{pang2021identifying}. Learned models can estimate binary contact states for individual hand links during grasping, but do not recover precise contact locations or normals~\cite{zhang2025robustdexgrasp}. 

Taken together, real-time force regulation over evolving whole-hand contacts remains largely unexplored on physical robotic systems. We address this gap by turning the hand itself into a geometric contact sensor. From joint proprioception, the hand kinematics, and a tracked object mesh, our method infers contacts on any hand link and places this geometric contact estimation directly inside the control loop—without tactile sensing. At every control cycle, previously unplanned contacts can enter or leave the contact set, and forces are reallocated subject to friction, actuator limits, and an actuation-consistency constraint motivated by whole-limb force analysis~\cite{bicchi1994problem}. We systematically evaluate robustness to constant object-pose offsets in simulation.

Extensive simulations show that real-time force reallocation maintains grasps under dynamic perturbations that cause non-adaptive execution to fail. 
Real-world experiments confirm stable grasp on a 27-DoF arm-hand system under human-applied disturbances.
Integrated with reactive reaching~\cite{lee2025hierarchical}, the regulator forms a fully reactive pipeline from grasp generation and acquisition to stabilization and recovery. To the best of our knowledge, this is the first hardware system to regulate grasp forces over a dynamically evolving set of whole-hand contacts (Fig.~\ref{fig:realworld_overview}).

\section{Related Work}

\subsection{Dexterous Grasp Synthesis}
Dexterous grasp synthesis has advanced rapidly, from force-closure optimization~\cite{wang2022dexgraspnet,zurbrugg2025graspqp,chen2025dexonomy} to generative models learned from the resulting large-scale datasets~\cite{weng2024dexdiffuser,zhang2024dexgraspnet}, producing diverse grasps across objects and hand embodiments.
Executing these grasps requires collision-free acquisition and stable interaction. 
Both precomputed contact forces and heuristic squeezing strategies~\cite{chen2025dexonomy, zurbrugg2025graspqp} can become unsuitable as contacts change, motivating online contact estimation and force allocation.

\subsection{Contact State Estimation}
Whole-hand force regulation needs the position and normal of every contact on the hand, and existing approaches each provide only part of this.
Tactile sensors give both directly where they are mounted, and tactile arrays now reach the palm and finger surfaces~\cite{zhao2025embedding,sharma2025self,zhang2025soft}.
Wiring and data management, however, still limit full coverage of an articulated hand~\cite{yousef2011tactile,slepyan2026scalable}.
Proprioception infers contact from joint-torque residuals without dedicated sensors~\cite{manuelli2016localizing,sipos2025multiscope}, but the residuals become ambiguous under multiple contacts~\cite{pang2021identifying}, which is the usual case in whole-hand grasps.
A learned estimator can reconstruct per-link contact from joint-state and action histories, supervised in simulation by the contact flags and impulses available to a privileged teacher~\cite{zhang2025robustdexgrasp}.
This covers every link, but yields only whether a link is in contact and how hard, not where or along which normal.
Vision combined with proprioception predicts contact without tactile sensing, but only at a fixed set of fingertip sites~\cite{patil2026nocontactnoworries}.

When the object mesh and pose are known, contact can instead be computed from geometry, which gives positions and normals on every link without any contact sensing.
This is common in human hand--object modeling, where contact is defined by the distance between the hand mesh and the object surface~\cite{taheri2020grab,grady2021contactopt}.
There the hand pose is itself an estimate, whereas a robot hand measures its joint angles, so only the object pose has to be estimated.
We adopt geometric contact estimation inside a real-time control loop.

\subsection{Grasp Force Regulation}
Allocating forces across a set of contacts under friction and actuation limits is a classical problem in dexterous grasping~\cite{kerr1986analysis}.
Whole-limb analyses extend it to contacts on any link and separate joint forces that can be actively controlled from passive mechanics and compliance~\cite{bicchi1994problem,haas2018passive}, while taking the contact configuration as given.
Reactive grasping on physical hands closes the loop at execution time, through object-level impedance control with passivity guarantees~\cite{wimboeck2006passivity}, adaptive finger coordination under disturbances and unknown dynamics~\cite{khadivar2023adaptive}, model predictive interaction control that maintains force closure~\cite{gold2021model}, and tactile-driven control of pinching~\cite{psomopoulou2021robust} and, more recently, of compliant and coordinated dexterous grasping~\cite{ke2026tacdexgrasp,yu2026coorgrasp}.
CoorGrasp~\cite{yu2026coorgrasp} allows general contact numbers and locations and reallocates forces online, provided tactile measurements are available at those contacts. Our distinction is the use of geometry and proprioception to update contacts across uninstrumented hand surfaces and include them in the allocation. This trades dedicated contact sensing for a known object model and tracked pose, and we demonstrate the resulting loop on a physical whole-hand system.

\section{Methodology}
\label{sec:methodology}

\subsection{System Overview and Problem Formulation}
\label{subsec:system_overview}

Our goal is to regulate contact forces in real time as the hand--object contact configuration evolves.
We consider a dexterous hand with $n_h$ actuated degrees of freedom (DoF) and joint configuration $q \in \mathbb{R}^{n_h}$, together with a target object whose 6-DoF pose is continuously tracked.
The object mesh $\mathcal{M}$ is obtained from a CAD model in simulation or from an offline reconstruction of the physical object. %

At each control cycle, the controller estimates contacts over the whole-hand surface and recomputes the force distribution; no contact forces are measured.
Specifically, we estimate the contact state $\mathcal{C}_t$ at time $t$ from $N_c(t)$ points sampled from $\mathcal{M}$ (Sec.~\ref{subsec:contact_estimation}).
We omit the time index below for clarity.
Then we recompute the force distribution on the points $f_c=[f_1^T,\dots,f_{N_c}^T]^T \in \mathbb{R}^{3N_c}$, and geometrically infer the net contact wrench on the object $w_c$ and the mapping to torque of actuated joints $\tau_q$ (Sec.~\ref{subsec:force_regulation}).
When the object moves and the contacts change, the forces ($f_c$, $w_c$, and $\tau_q$) are redistributed over the new contact configuration $\mathcal{C}_t$.
Instead of estimating and compensating for gravity, inertial loads, or external wrenches, we adopt a quasi-static formulation and seek a squeezing force distribution that resists unmodeled loads through friction (Sec.~\ref{subsec:grasp_reaching}).

\subsection{Geometric Whole-Hand Contact Estimation}
\label{subsec:contact_estimation}

We treat contact detection as a geometry problem rather than a sensing problem.
Given the object's mesh and tracked pose and the hand's measured joint angles, the distance from any point on the hand surface to the object is fully determined, so contacts on every link follow without any contact sensor.
For the object mesh $\mathcal{M}$, we precompute a signed distance field (SDF) in its canonical frame.
For each link, we draw 2000 points uniformly on its mesh and reduce them by farthest-point sampling to a density of 2 points per cm$^2$.
This is done once and gives about 3000 candidate points over the whole hand in both simulation and hardware.
At each control cycle, the candidate points are transformed into the object frame using the current object pose and joint configuration $q$, and their signed distances are queried from the SDF.
Points with distance below $d_{\mathrm{thresh}}$ are contact candidates, and for each link we keep up to $K$ of them closest to the object surface.
The points define $\mathcal{C}_t$
\begin{equation}
\mathcal{C}_t = \{(p_i,n_i,\ell_i)\}_{i=1}^{N_c(t)},
\end{equation}
where $p_i$, $n_i$, and $\ell_i$ denote the contact position, object surface normal, and hand link, respectively.
Because every link surface is checked at every cycle, new contacts can appear on any part of the hand as the object moves within the grasp.
The SDF queries run in parallel with NVIDIA Warp~\cite{Macklin_Warp_A_High-performance_2022}, so the whole-hand search takes under $1\,\mathrm{ms}$.

\begin{figure*}[t]
    \centering
    \includegraphics[width=1.0\linewidth]{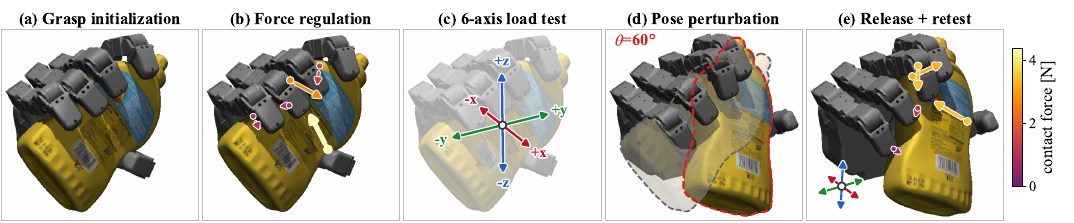}
    \caption{Visualization of perturbation-test protocol.
    Starting from a stored grasp, the controller regulates the contact forces~(a,~b), and a force equal to 10\% of the object weight is applied along one of $\pm x,\pm y,\pm z$ at a time~(c).
    A spring--damper wrench then pulls the object toward a pose rotated by $\theta$ about a random axis~(d; grey/red dashed outline: before/after), and the load test is repeated~(e).
    A trial succeeds only if the object is held through all six directions.
    }
    \label{fig:experiment_setup}
\end{figure*}

\subsection{Real-Time Whole-Hand Force Allocation}
\label{subsec:force_regulation}

At every control cycle the forces are solved over the current contact set, palm included, so the distribution changes whenever the contacts do.
Let $f_c=[f_1^T,\dots,f_{N_c}^T]^T \in \mathbb{R}^{3N_c}$ denote the contact forces, each expressed in its local contact frame.
The net contact wrench on the object is
\begin{equation}
w_c = G f_c,
\label{eq:grasp_matrix}
\end{equation}
where the grasp matrix $G \in \mathbb{R}^{6\times 3N_c}$ is constructed from the current contact positions and normals.

We recompute the force distribution from the contacts that exist now, not from the grasp that was planned.
We pose this allocation as a convex quadratic program (QP) whose objective minimizes the residual object wrench:
\begin{equation}
\min_{f_c} \; \frac{1}{2}\|Gf_c\|^2.
\label{eq:qp_objective}
\end{equation}
The implementation uses unit numerical weights on force (N) and torque (N$\cdot$m) components. The optimization is subject to friction, minimum grasp force, actuator limits, and actuation-consistency constraints.

\subsubsection{Contact Constraints}
For each contact, let $f_i=[f_{i,n},f_{i,t1},f_{i,t2}]^T$ be expressed in its local contact frame.
We impose unilateral contact and a linearized friction constraint~\cite{stewart1996implicit,anitescu1997formulating}:
\begin{equation}
f_{i,n} \ge 0,
\label{eq:unilateral}
\end{equation}
\begin{equation}
|f_{i,t1}| + |f_{i,t2}| \le \mu f_{i,n},
\quad \forall i,
\label{eq:friction}
\end{equation}
where $\mu$ is the static friction coefficient and Eq.~(\ref{eq:friction}) is the pyramid inscribed in the Coulomb cone, i.e., a conservative approximation.
To prevent the trivial zero-force solution, we additionally impose
\begin{equation}
\sum_{i=1}^{N_c} f_{i,n} \ge f_{n,\mathrm{min}}.
\label{eq:min_normal_force}
\end{equation}
The QP seeks a squeezing distribution with a small residual wrench; exact balance is obtained only when compatible with the constraints. This allocation provides frictional resistance to unmodeled loads rather than explicitly compensating for them.
The level $f_{n,\min}$ sets a lower bound on the total normal force; we use $f_{n,\min}=10$\,N, at the upper end of the contact-force range reported for synthesized grasps~\cite{chen2025dexonomy}.

\subsubsection{Actuator Torque Limits}
Let $N_{\mathrm{act}} \le N_c$ denote the number of contacts located on actuated links, and let $f_{\mathrm{act}} \in \mathbb{R}^{3N_{\mathrm{act}}}$ collect their corresponding force vectors.
For these actuated contacts, the mapping to joint torques is
\begin{equation}
\tau_q = J_{\mathrm{act}}(q)^T f_{\mathrm{act}},
\label{eq:force_torque_mapping}
\end{equation}
where $J_{\mathrm{act}} \in \mathbb{R}^{3N_{\mathrm{act}}\times n_h}$ stacks the corresponding contact Jacobians.
Unactuated contacts, such as those on the palm, exist in $w_c$ but not $f_{\mathrm{act}}$ or $J_{\mathrm{act}}$; their forces are treated as passive reactions.

For contacts on actuated finger links, the required joint torques must satisfy
\begin{equation}
-\tau_{\mathrm{max}}
\le
J_{\mathrm{act}}(q)^T f_{\mathrm{act}}
\le
\tau_{\mathrm{max}},
\label{eq:torque_limit}
\end{equation}
where $\tau_{\mathrm{max}} \in \mathbb{R}^{n_h}$ denotes the hand joint torque limits.

\subsubsection{Actuation-Consistency Constraint}
Torque limits alone do not prevent the optimizer from assigning force components that are invisible to the controlled joint torques.
Classical whole-limb analyses distinguish actively controllable contact forces from components determined by passive mechanics and compliance~\cite{bicchi1993force,bicchi1994problem,prattichizzo1998dynamic}.
Rather than explicitly modeling these compliance-dependent effects, we use a conservative actuation-consistency constraint that suppresses desired force components in $\mathrm{Null}(J_{\mathrm{act}}^T)$.

We compute the full SVD $J_\mathrm{act}=U\Sigma V^T$ and let $N_u$ collect the columns of $U$ beyond $\mathrm{rank}(J_\mathrm{act})$, determined with a singular-value tolerance of $10^{-6}$, so that its columns span $\mathrm{Null}(J_\mathrm{act}^T)$.
We then impose
\begin{equation}
-\epsilon_{\mathrm{margin}}\mathbf{1}
\le
N_u^T f_{\mathrm{act}}
\le
\epsilon_{\mathrm{margin}}\mathbf{1}.
\label{eq:null_space_constraint}
\end{equation}
This bounds the desired force components in the chosen SVD null-space basis, reducing reliance on forces invisible to finger-joint torques. It is a conservative allocation restriction, not a guarantee that the physical contacts realize the desired forces; their realization also depends on contact compatibility, compliance, and passive reactions~\cite{haas2018passive}.

At every control cycle, $\mathcal{C}_t$, $G$, and $J_{\mathrm{act}}$ are updated and the QP is solved again.
The optimized finger-contact forces are converted to joint torque commands as
\begin{equation}
\tau_q^* = J_{\mathrm{act}}(q)^T f_{\mathrm{act}}^*,
\end{equation}
forming the real-time contact-estimation and force-regulation loop. 
If the solve fails, the controller retains the last feasible QP torque command, initialized to zero.

\subsection{Integration with Reactive Reaching}
\label{subsec:grasp_reaching}

Force regulation operates only after the hand has reached the grasp pose.
Adding a reactive reaching controller in front makes the whole sequence autonomous, from approach to hold.
We extend the hierarchical reactive reaching controller of Lee et al.~\cite{lee2025hierarchical} to our five-finger arm--hand system.
Target grasps come from Dexonomy~\cite{chen2025dexonomy}, each paired with a pre-grasp that keeps at least $3\,\mathrm{cm}$ of clearance from the object for a collision-free approach, computed with the penetration-depth metric of GraspQP~\cite{zurbrugg2025graspqp}.
We select the pre-grasp closest to the current palm pose, and both targets follow the tracked object pose so that the hand reacts to object motion before contact.

During reaching, we jointly regulate the poses of all five fingertips and the palm.
Let $v_i^d$ and $\omega_i^d$ denote the desired linear and angular velocities of fingertip $i$, and let $v_p^d$ and $\omega_p^d$ denote those of the palm.
We compute the joint velocity command by solving
\begin{equation}
\begin{aligned}
\dot q^* = \arg\min_{\dot q}\;&
\sum_{i=1}^{5}
\left(
w_x \|J_{x_i}\dot q-v_i^d\|^2
+
w_R \|J_{R_i}\dot q-\omega_i^d\|^2
\right) \\
&+
w_{p,x}\|J_{x_p}\dot q-v_p^d\|^2
+
w_{p,R}\|J_{R_p}\dot q-\omega_p^d\|^2 ,
\end{aligned}
\label{eq:reactive_reaching}
\end{equation}
subject to linearized self-collision, environment-collision, and hand--object collision constraints.
Here $J_{x_i}$ and $J_{R_i}$ are the positional and rotational Jacobians of fingertip $i$, $J_{x_p}$ and $J_{R_p}$ those of the palm, and $w$ are task weights.
A configuration regularizer keeps the hand near a nominal posture.
The fingertip task weights increase as the hand approaches the target, while the palm task weights remain fixed throughout the motion.

Stage transitions are triggered by geometry.
When the mean fingertip error to the pre-grasp falls below $2\,\mathrm{cm}$, the target switches to the grasp pose and the hand--object collision constraints are released.
When the thumb and two other fingers are in contact, control hands over to force regulation.
If the convex hull of the fingertips and palm no longer intersects the object's bounding box, the grasp is lost and control returns to reaching.

\section{Experiments and Results}
\label{sec:experiments}
We evaluate the framework in simulation, where controlled perturbations allow quantitative comparison and ablation (Secs.~\ref{subsec:sim_studies} and~\ref{subsec:ablation}), and on a physical arm--hand system under human disturbances (Sec.~\ref{subsec:real_world_exp}).
Unless otherwise noted, we use $\mu=0.7$, $f_{n,\min}=10$\,N, $\tau_{\max}=0.2$\,N$\cdot$m per joint, $d_\mathrm{thresh}=1$\,cm, $K=5$, and $\epsilon_\mathrm{margin}=0.5$\,N. The QP is solved with OSQP~\cite{stellato2020osqp}.

\subsection{Simulation Studies}
\label{subsec:sim_studies}
We evaluate in MuJoCo~\cite{todorov2012mujoco} on the 79 YCB objects with a released laser-scan mesh, each with 10 power grasps from Dexonomy~\cite{chen2025dexonomy} (five Large Diameter, five Sphere 4-Finger), chosen because they engage the whole hand.
Gravity is disabled to isolate grasp reliability from gravitational loading, so an object may remain inside the hand even when it is not securely held.
We therefore apply a force of $0.1\,mg$ (10\% of the object weight) for 1\,s along each of $\pm x,\pm y,\pm z$ in turn.
A trial fails if the grasp-loss criterion in Sec.~\ref{subsec:grasp_reaching} is met at any stage.
All 790 grasps are used for the contact-count study  (Sec.~\ref{subsubsec:contacts}).
The perturbation study (Sec.~\ref{subsubsec:perturbation}) needs repeated trials over random axes, so it uses five representative objects selected by $k$-means over bounding-box extent, aspect ratio, and convexity. 
These are the mustard bottle, extra-large clamp, cup (065-f), Lego Duplo (073-f), and Rubik's cube.

\begin{figure*}[t]
    \centering
    \includegraphics[width=1.0\linewidth]{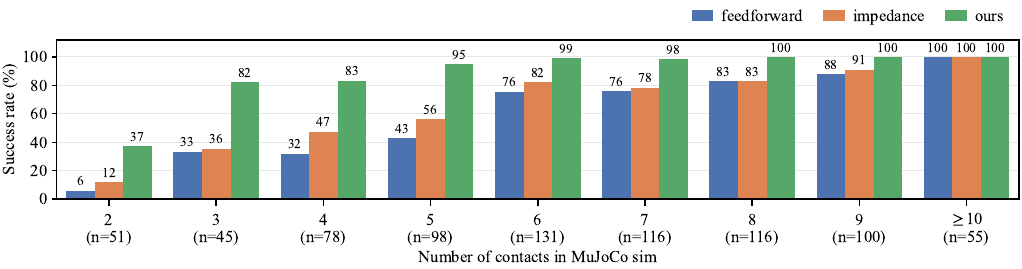}
    \caption{
Success rate versus number of hand--object contacts in MuJoCo simulation.
790 power grasps on 79 YCB objects under the six-direction load test of Fig.~\ref{fig:experiment_setup}(a--c), binned by the number of ground-truth contacts at initialization; $n$ is the number of grasps per bin.}
    \label{fig:success_vs_contact}
\end{figure*}

\begin{figure}[t]
    \centering
    \includegraphics[width=1.0\linewidth]{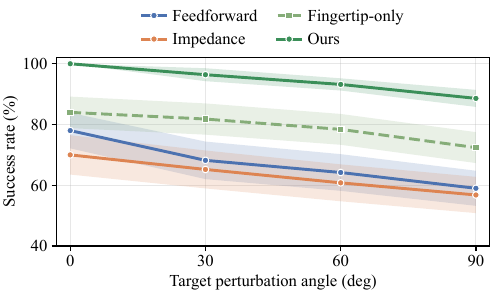}
    \caption{Grasp robustness under perturbations across five YCB objects (50 grasps per method). At $0^\circ$: initial six-direction load test; at $30^\circ$--$90^\circ$: 10 trials per grasp, including initial failures. Bands: $\pm1$ standard error across grasp-level rates.}
    \label{fig:success_vs_perturbation}
\end{figure}

\subsubsection{Dynamic Perturbation Resistance}
\label{subsubsec:perturbation}
We test whether recomputing forces from the current contacts keeps the grasp when the object moves within the hand.
A virtual spring--damper creates this motion by pulling the object toward a pose rotated by $0^\circ$--$90^\circ$ about a random axis (Fig.~\ref{fig:experiment_setup}).
We compare against two non-adaptive baselines that follow the standard execution pipeline of grasp synthesis methods.
Both allocate contact forces once at grasp initialization from the ground-truth MuJoCo contacts, following Dexonomy~\cite{chen2025dexonomy}, and differ only in how the allocated forces are applied.
\emph{Feedforward torque control} maps the forces to joint torques through the contact Jacobians and holds these torques for the remainder of the trial.
\emph{Task-space impedance control} instead offsets each contact's target position along the allocated force, so that a Cartesian impedance controller at the initial configuration produces that force.
These targets remain fixed when the object moves. 
\emph{Fingertip-only} runs our controller with contact candidates restricted to the five distal links, isolating contact coverage from online adaptation.

Ours outperforms both non-adaptive baselines at every perturbation angle (Fig.~\ref{fig:success_vs_perturbation}), showing that forces allocated once at initialization do not survive object motion.
Whole-hand contacts also beat Fingertip-only at every angle, $88.6\%$ versus $72.4\%$ at $90^\circ$.
Contact coverage thus matters both for settling the initial grasp ($0^\circ$) and for holding it under disturbance.

\subsubsection{Impact of Multi-Link Whole-Hand Contacts}
\label{subsubsec:contacts}
We test how the number of contacts a grasp makes affects success, using all 790 grasps under the six-direction load test.
Although all are power grasps, the number of contacts realized at initialization ranges from 2 to more than 10 depending on the hand--object configuration, and about a third have five or fewer.
Fig.~\ref{fig:success_vs_contact} reports success as a function of the ground-truth MuJoCo contact count $N_c^{\mathrm{gt}}$, over which the baselines allocate their forces; our method uses only its own geometric estimate.
Most failures occur while the grasp settles, few during the load test.
Since the allocated forces cannot be realized exactly, the object moves slightly while the grasp forms and the contact locations change.
The baselines keep applying forces for the initial contacts and lose the object, whereas our method updates the contacts and forces as they change and lets the grasp settle.
The gap is largest at $N_c^{\mathrm{gt}}=3$--$5$ (32--56\% versus 82--95\%) and closes at $N_c^{\mathrm{gt}}\ge 10$.
Success also increases with $N_c^{\mathrm{gt}}$ for all methods, presumably because more contacts enlarge the set of wrenches the grasp can resist.

\subsection{Ablation Studies}
\label{subsec:ablation}

\subsubsection{Effect of Actuation-Consistency Constraint}

\begin{figure}[t]
    \centering
    \includegraphics[width=1.0\linewidth]{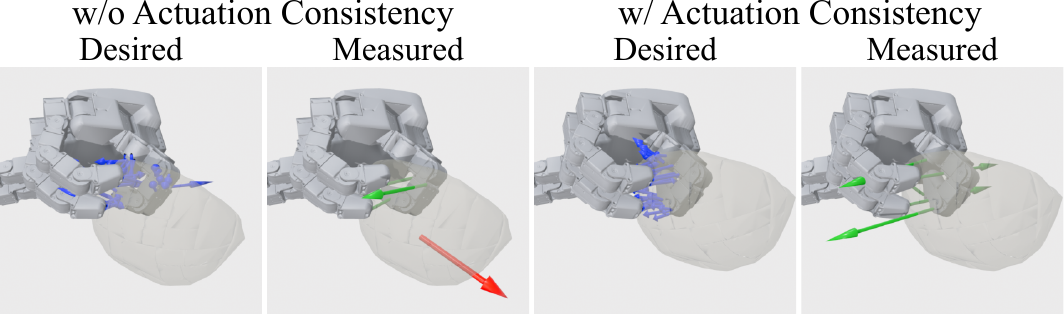}
    \caption{
Effect of the actuation-consistency constraint on a three-finger grasp, 2\,s after initialization.
Blue: desired contact forces; green: measured contact forces; red: measured net contact force at the COM, nearly zero with the constraint.
Blue arrows use twice the length scale of green arrows.
}
    \label{fig:feasible_force}
\end{figure}

We test whether the actuation-consistency constraint, which suppresses desired force components in $\mathrm{Null}(J_\mathrm{act}^T)$, changes the forces the hand actually applies.
In the example of Fig.~\ref{fig:feasible_force}, enforcing the constraint reduces the measured net contact force on the object to nearly zero.
Without it, the measured contact forces leave a resultant of $0.428\,\mathrm{N}$ (red arrow), because the allocation counts on components the joints cannot produce.

To evaluate the effect across grasp types, we synthesize power grasps (Type~1) and three-finger precision grasps (Type~7) with Dexonomy~\cite{chen2025dexonomy}, giving $96$ power and $68$ precision grasps after quality filtering.
Under the six-direction load test, power grasps succeed $100\%$ with or without the constraint, whereas precision-grasp success increases slightly from $38.5\%$ to $42.1\%$ with it.
The low absolute rate of precision grasps reflects their sparse contacts: with fewer contact points there is less margin against slip, consistent with the contact-count trend in Sec.~\ref{subsubsec:contacts}.

\subsubsection{Robustness to Object Pose Error}
We test how sensitive the controller is to object pose error, since real tracking is never perfect.
We inject constant synthetic rotational ($0^\circ$--$30^\circ$) and translational ($0$--$5\,\text{mm}$) offsets into the object pose provided to the controller.
We compare our real-time force regulation across three contact thresholds ($d_{\text{thresh}}=0.5,1,2\,\text{cm}$) against a non-adaptive feedforward baseline, with five random offset directions per grasp at each perturbation level.
For the feedforward baseline, the forces allocated at initialization are transformed from the object’s CAD frame to the world frame using the perturbed object pose.

\begin{figure}[t]
    \centering
    \includegraphics[width=1.0\linewidth]{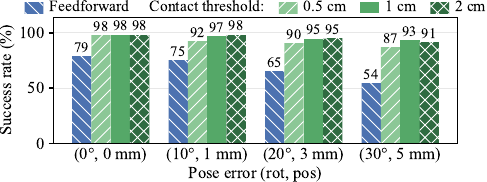}
    \caption{Robustness to synthetic object pose error.
While the non-adaptive feedforward baseline degrades with increasing pose error, real-time contact and force adaptation maintains high grasp success.}
    \label{fig:contact_thres_ablation}
\end{figure}

As shown in Fig.~\ref{fig:contact_thres_ablation}, feedforward performance drops from $79\%$ to $54\%$ as the synthetic pose offset increases, whereas real-time force regulation maintains $87$--$93\%$ success even at the largest one.
This supports robustness to constant pose bias; time-varying noise, latency, and tracking loss are not evaluated by this test.
Among the tested thresholds, $d_{\text{thresh}}=1\,\text{cm}$ provides the best robustness, achieving $93\%$ success at $(30^\circ,5\,\text{mm})$.

\subsubsection{Effect of Max Number of Contacts per Link}

\begin{figure}[t]
    \centering
    \includegraphics[width=1.0\linewidth]{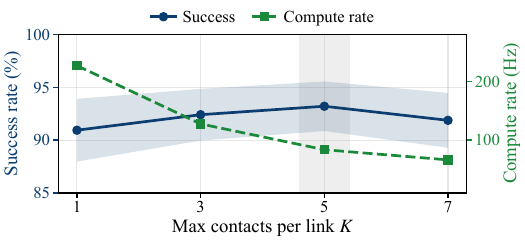}
    \caption{
Effect of the maximum number of contacts per link $K$.
750 attempts per $K$: 50 grasps, three perturbation angles, five repeats; initial failures included. Band: $\pm1$ standard error across grasp means. Compute rate is reciprocal mean compute time; shading marks $K=5$.
}
    \label{fig:mcpl_ablation}
\end{figure}

We test how many contact points per link the QP should receive, since real-time force regulation depends on two things that pull in opposite directions: how fast the loop runs and how well each contact is represented.
In practice a contact is a small patch rather than a single point.
Passing several nearby points per link lets the QP spread force over the patch rather than load a single point, which should make the allocation less sensitive to small errors in contact location.
But every added point brings its own force variables and constraints, so the QP slows down and the controller reacts later to contact changes.
We sweep $K\in\{1,3,5,7\}$ on the five objects and grasps of Sec.~\ref{subsubsec:perturbation}, including initial settling failures (Fig.~\ref{fig:mcpl_ablation}).
Success rises from $90.9\%$ at $K=1$ to $93.2\%$ at $K=5$ and falls back to $91.9\%$ at $K=7$, while the mean QP contact count grows from $14$ to $91$ and the control rate drops from $227$ to $66$\,Hz.
The trend fits the two effects: a few points per link are enough to capture the patch, and beyond that the slower loop costs more than the extra points return.
We use $K=5$, which gives the highest success.

\subsection{Real World Experiments}
\label{subsec:real_world_exp}

\begin{figure}[t!]
    \centering
    \includegraphics[width=1.0\linewidth]{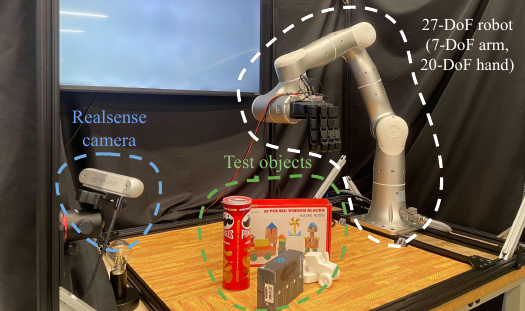}
    \caption{Real-world experimental setup.}
    \label{fig:realworld-setup}
\end{figure}

\begin{figure*}
    \centering
    \includegraphics[width=1.0\linewidth]{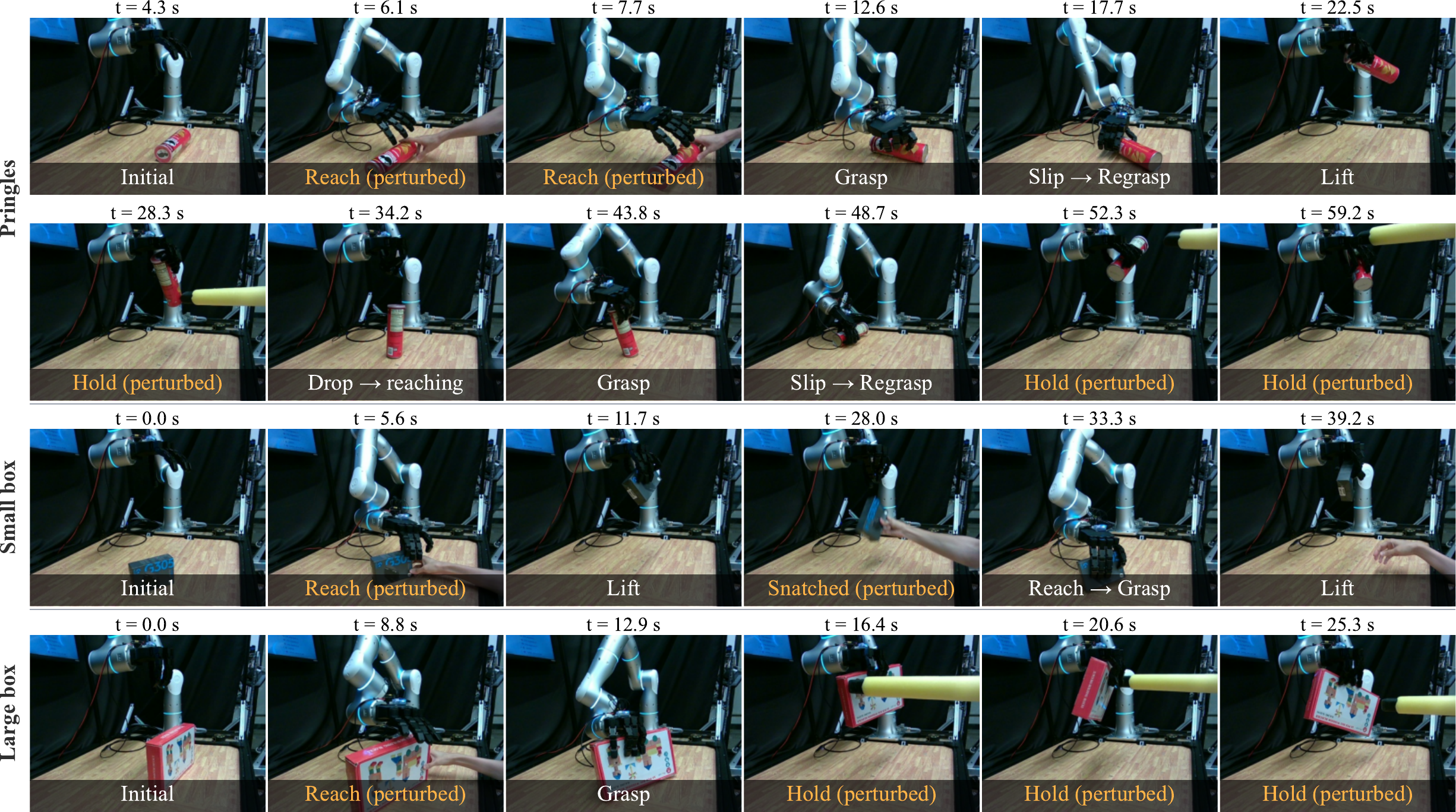}
    \caption{
Execution snapshots of the full pipeline on three objects, combining hierarchical reactive reaching~\cite{lee2025hierarchical} with the whole-hand force regulation.
Frames with human perturbation are labeled in orange.
The hand follows the object displaced during reaching, regulates the grasp under disturbances after contact, and, when the object slips or is snatched away, returns to reaching and regrasps it at its new pose.
Timestamps are in seconds.}
    \label{fig:realworld-reaching}
\end{figure*}

\subsubsection{Setup}
We deploy the system on a 27-DoF platform (a 7-DoF Flexiv Rizon 4 arm and a 20-DoF Robotis 5F hand) with an Intel RealSense D455 (Fig.~\ref{fig:realworld-setup}); the hand's fingertip tactile sensors are not used.
Contact estimation uses the same link sampling and $K=5$ as in simulation.
Because the tracked pose arrives with latency, contact estimation and the QP run once per tracked pose, using the joint configuration recorded at the image timestamp so that the pose and the hand configuration refer to the same instant.
Computations run on a desktop with an Intel i9-13900K and an NVIDIA GeForce RTX 4090.

\subsubsection{Object Reconstruction and Tracking}
Grasp generation and tracking need an object model, which we obtain from a short handheld scan with a single RGB-D camera.
Point2Pose~\cite{lin2026point2pose} tracks the camera pose during the scan, and 2D Gaussian Splatting~\cite{huang20242d} reconstructs the object.
We extract the mesh $\mathcal{M}$ from the Gaussians by Poisson surface reconstruction~\cite{kazhdan2006poisson} and precompute its SDF.
For tracking, we replace the mesh renderer in FoundationPose~\cite{wen2024foundationpose} with a Gaussian renderer and use the modified tracker to estimate the object's 6-DoF pose at 30\,Hz.

\begin{table}[t]
\centering
\caption{Summary of full-pipeline runs across three objects. Losses denote the hand losing the object, while unrecovered losses correspond to workspace exits.}
\label{tab:realworld}
\begin{tabular}{lcccc}
\toprule
Object & Duration [s] & Perturbations & Losses & Recovered \\
\midrule
Pringles  & 59 & 16 & 3 & 3/3 \\
Small box & 48 & 31 & 2 & 1/2 \\
Large box & 94 & 31 & 3 & 2/3 \\
\bottomrule
\end{tabular}
\end{table}

\subsubsection{Results}
Fig.~\ref{fig:realworld-reaching} and Table~\ref{tab:realworld} summarize three runs on three objects with human perturbations.
The perturbations target both stages of the pipeline: the object is displaced while the hand reaches for it, and pushed and twisted with a stick while the hand holds it.
The hand loses the object in eight of the 78 perturbations over about 200\,s.
In six of these the hand returns to reaching once the loss criterion of Sec.~\ref{subsec:grasp_reaching} fires and regrasps the object at its new pose without intervention (see supplementary video).
In the other two the object leaves the workspace; the robot then returns to a default pose for safety, and the run continues only after the object is repositioned by hand and the reaching controller restarted.
These two are limits of the arm's workspace rather than of the controller.

\begin{figure}[t!]
    \centering
    \includegraphics[width=1.0\linewidth]{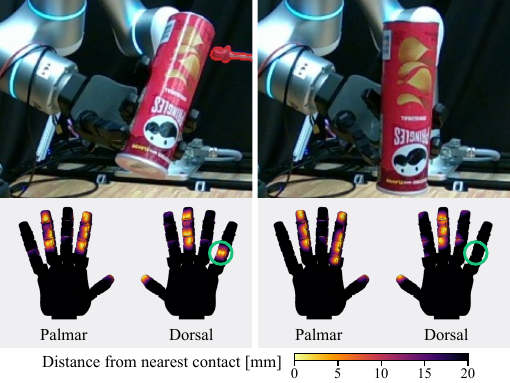}
    \caption{
A manually posed grasp with fingers alternating in front of and behind the object, regulated by the whole-hand controller.
Bottom: palmar and dorsal views colored by distance to the nearest contact.
Cyan ellipses show the change in estimated contact on the little finger's dorsal surface.}
    \label{fig:fun-grasp}
\end{figure}

Because contacts are searched over the whole hand surface, the estimator is not limited to grasps that a taxonomy or a sensor layout anticipates.
In Fig.~\ref{fig:fun-grasp} we manually pose the fingers alternating across the front and back of the object, so that some hold it with their dorsal surface.
No grasp-type prior would place a contact there, but to the estimator it is just another point on a link mesh, and the QP handles it through the same contact Jacobian.
The controller holds the grasp while the object is pushed with a rod, and the estimated dorsal contact on the little finger shifts with the object (cyan ellipses).

\addtolength{\textheight}{-0.5cm} %

\section{Discussion and Conclusion}
\label{sec:conclusion}
We present a whole-hand force regulation controller that needs no contact sensing.
Contacts are computed from the tracked object pose and the joint angles, and the forces on them are recomputed as the contacts change.
In simulation, this outperforms non-adaptive and fingertip-only baselines.
On hardware, combined with a reactive reaching controller, the system runs autonomously from approach to hold and regrasps after most losses.

Several limitations remain.
First, the controller needs an object model and a pose tracker; when tracking fails under heavy occlusion, the contact estimates fail with it.
Second, thin objects are hard: a pose error of a few millimeters can put sampled hand points on the far side of the object and flip the contact normal.
Third, the force allocation is quasi-static, with no explicit compensation for gravity, inertia, or external loads.
Finally, the actuation-consistency constraint omits passive compliance and contact preloads~\cite{bicchi1994problem,haas2018passive}.
Future work will address these dynamic and compliant effects and the robustness of contact estimation under degraded tracking.

\bibliographystyle{IEEEtran}
\bibliography{references}

\end{document}